\documentclass{article}

\usepackage{arxiv}

\usepackage[utf8]{inputenc} % allow utf-8 input
\usepackage[T1]{fontenc}    % use 8-bit T1 fonts
\usepackage{url}            % simple URL typesetting
\usepackage{booktabs}       % professional-quality tables
\usepackage{amsfonts}       % blackboard math symbols
\usepackage{nicefrac}       % compact symbols for 1/2, etc.
\usepackage{microtype}      % microtypography
\usepackage{lipsum}		% Can be removed after putting your text content
\usepackage{graphicx}
\usepackage{natbib}
\usepackage{doi}

\usepackage[justification=centering]{caption}
\usepackage{leftidx}
\usepackage{amsmath}
\usepackage{makecell}
\usepackage{array}

\usepackage{subfigure}
\usepackage{amssymb}

\usepackage{glossaries}

\makeglossaries
\setglossarypreamble{To illustrate the terminologies used in this report, a CAD model example is shown in Fig.~\ref{Fig.6}. The model is based on a block. One of its corners is removed and a smaller block is added in the center.}

\newglossaryentry{edge cluster}{%
  name={edge cluster},
  description={An edge cluster is defined as a group of edges that share the same convexity characteristic and are adjacent to each other in the vertex-edge graph of a TAAG. The number of edges within one group may be one or more. As shown in Fig.7(b), there are three groups of edges: $EG_1$: $\{E_1, E_2, E_3\}$, $EG_2$: $\{E_4, E_5, E_6, E_7, E_8, E_9, E_{10}, E_{11}, E_{12}\}$, $EG_3$: $\{E_{13}, E_{14}, E_{15}, E_{16} \ldots \}$. Within each edge cluster, the edges are connected by vertices. According to the convexity of edges, we can get convex edge groups $EG_1$, $EG_3$, and a concave one $EG_2$.}
}

\newglossaryentry{leaf edge}{%
  name={leaf edge},
  description={A leaf edge means that its two ends can only be connected by a unique path within an edge cluster. In other words, if two ends of an edge can be connected by more than one path, the edge cluster has a loop inside. In $EG_2$, $\{E_4, E_5, E_6, E_7, E_8, E_9\}$ constitutes a loop while \{E$_{10}$, E$_{11}$, E$_{12}$\} are leaf edges.}
}

\newglossaryentry{edge loop}{%
  name={edge loop},
  description={An edge loop is a set of edges that constitute a loop. An edge loop contains no leaf edges. This concept only refers to the whole loop rather than individual edges. For example, the edges $\{E_4, E_5, E_6, E_7, E_8, E_9\}$ in $EG_2$ constitute an edge loop $EL_1$.}
}

\newglossaryentry{feature subgraph}{%
  name={feature subgraph},
  description={A feature subgraph is defined as a set of connected faces and edges that have the same convexity classes or are transitory in the edge-face graph of a TAAG. If all faces and edges are convex or transitory, this feature subgraph is convex. If faces and edges are concave or transitory, this feature subgraph is concave. For example, $\{F_1, F_2, F_3, E_1, E_2, E_3\}$ forms a convex feature area and $\{F_4, F_5, F_6, E_{10}, E_{11}, E_{12}\}$ form a concave feature area.}
}

\newglossaryentry{division edge}{%
  name={division edge},
  description={If the convexity of an edge is different from that of its neighboring geometric entities in an edge-face graph, this edge is called a division edge. For example, $E_4$ is concave. There are many convex edges, such as $E_1$ and $E_{13}$, on the two sides of $E_4$. So $E_4$ is a division edge.}
}

\newglossaryentry{feature boundary}{%
  name={feature boundary},
  description={A feature boundary consists of a set of division edges that are connected as an edge cluster. That is, there is a path between any two division edges of a feature boundary.},
  plural = {feature boundaries}
}

\title{A Set-based Approach for Feature Extraction of 3D CAD Models}

\date{} 					% Or removing it

\author{Peng Xu\\
	School of Mechanical Engineering\\
	Shandong University\\
	Jinan, Shandong 250061 \\
	\texttt{xupeng0.618@gmail.com} \\
	\And
	Qi Gao \\
	School of Mechanical Engineering\\
	Shandong University\\
	Jinan, Shandong 250061 \\
	\texttt{gaoqi@sdu.edu.cn} \\
	\AND
	Ying-Jie Wu \\
	School of Mechanical Engineering \\
    Shandong University \\
    Jinan, Shandong 250061 \\
	\texttt{wyjcfy@qq.com} \\
}

\renewcommand{\headeright}{Technical Report}
\renewcommand{\undertitle}{Technical Report}

\begin{document}

\maketitle

\begin{abstract}
Feature extraction is a critical technology to realize the automatic transmission of feature information throughout product life cycles. As CAD models primarily capture the 3D geometry of products, feature extraction heavily relies on geometric information. However, existing feature extraction methods often yield inaccurate outcomes due to the diverse interpretations of geometric information. This report presents a set-based feature extraction approach to address this uncertainty issue. Unlike existing methods that seek accurate feature results, our approach aims to transform the uncertainty of geometric information into a set of feature subgraphs. First, we define the convexity of basic geometric entities and introduce the concept of two-level attributed adjacency graphs. Second, a feature extraction workflow is designed to determine feature boundaries and identify feature subgraphs from CAD models. This set of feature subgraphs can be used for further feature recognition. A feature extraction system is programmed using C++ and UG/Open to demonstrate the feasibility of our proposed approach.
\end{abstract}

% keywords can be removed

\section{Introduction}
\label{intro}
Computer-aided technology (CAx) plays a pivotal role in modern product lifecycle management (PLM) by revolutionizing the way companies design, develop, manufacture, and support products. By leveraging computer-aided design (CAD), computer-aided manufacturing (CAM), and other related technologies, companies can streamline and optimize various processes across the product lifecycle (PLC). CAD software enables designers to create detailed digital representations of products, facilitating rapid prototyping, visualization, and virtual testing. CAM systems automate manufacturing processes, increasing efficiency, precision, and scalability. How to integrate these CAxes becomes the central topic of PLM. 

Feature-based modeling is believed to bridge the gap among various CAxes as it allows for smoother collaboration and data exchange during the PLC~\citep{besharati2020fundamentals}. According to Zhang and Artling~\citep{zhang1994manufacturing}, features are the generic shapes with which design and manufacturing engineers associate attributes and knowledge in reasoning about products. Common types of features include geometric features, precision features, material features, assembly features, and technological features ~\citep{shah1990asu}. Among them, a geometric feature refers to a geometric shape or configuration of a part or component. This report focuses on geometric features as they directly represent the geometric attributes of CAD models. However, many existing CAD models do not contain geometric feature information inside. Thus, how to identify geometric features automatically is essential to realize feature-based modeling.

\begin{figure*}
  \centering
  \includegraphics[width=0.75\textwidth]{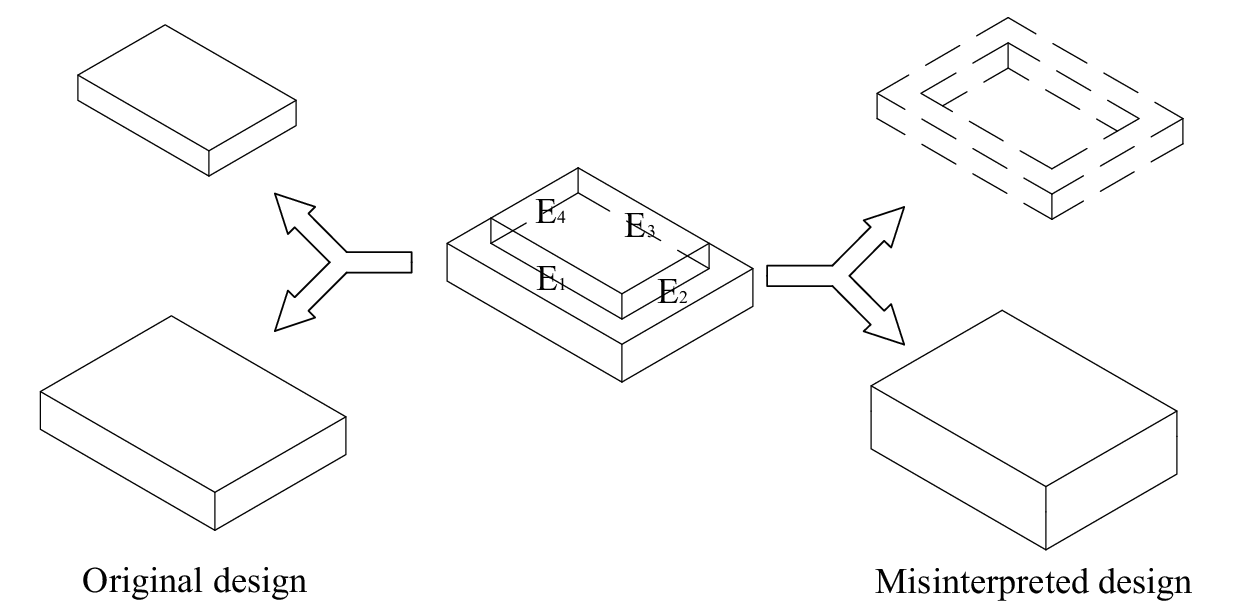}
  \caption{Two interpretations of a CAD model}
\label{fig:3}       % Give a unique label
\end{figure*}

Until now, many feature recognition methods have been developed to recognize geometric features automatically, such as design-by-feature (DBF) and automatic feature recognition (AFR). DBF requires building a geometric feature library first according to manufacturing guidelines. This library limits the application of DBF as it is not always available. AFR is identified as the most promising method because it can significantly reduce recognition efforts and provide consistent results across different CAD models and platforms. AFR methods include graph-based, volume decomposition-based, hint-based, and hybrid approaches ~\citep{shah2001discourse,babic2008review}. All these methods have two stages ~\citep{babic2008review}. The first stage is called feature extraction which characterizes geometric entities into geometric features. The second stage is feature recognition which matches the extracted geometric feature with feature patterns based on certain rules. However, current AFR methods may generate inaccurate geometric features from CAD models because they only rely on geometric information while other important features, such as precision features, are ignored in the recognition process. The reason is geometric information may be interpreted differently when it is inadequate to form geometric features. For example, the CAD model in Fig.~\ref{fig:3} is originally designed as a combination of two blocks. But it can also be interpreted as a block minus a circular step and the circular step is formed by connecting edges \{$E_1$, $E_2$, $E_3$, $E_4$\} and their neighboring planes. Which interpretation is correct depends on other design features.

This report aims to solve this uncertainty issue by proposing a set-based feature extraction approach to identify \glspl{feature subgraph} from CAD models. Our approach focuses on extracting \glspl{feature boundary} of geometric models and identifying all feature subgraphs from CAD models for further feature recognition. The reason is feature boundaries are invariant under different interpretations. All uncertainty information is preserved in the extraction results of our approach, and recognition decisions will not be made until more features, such as precision features or assembly features, are collected for feature recognition. This approach is inspired by set-based design (SBD) to deal with the uncertainty issue. SBD works simultaneously with a plethora of options, instead of converging quickly to a single option ~\citep{bernstein1998design, xu2023concept, xu2022parallel}. The remainder of this paper is structured as follows. Section ~\ref{Literature review} reviews the background of feature extraction. Section ~\ref{Methodology} presents the proposed set-based approach, including the definition of convexity of geometric entities, a two-level attributed adjacency graph (TAAG) model, and feature extraction algorithms. A feature extraction application system is developed to demonstrate our approach in Section ~\ref{Implementation}. A summary of the conclusions and future directions is presented in Section ~\ref{Conclusion}. Finally, a list of important terminologies is provided after the conclusions.

\section{Background}
\label{Literature review}
Many previous research has addressed the extraction of geometric features from 3D CAD models. In 1987, Joshi \citep{joshi1987cad} implemented the concept of attributed adjacent graph (AAG). An AAG is an edge-face graph of a solid model with attribute values assigning the arcs. Feature extraction was implemented by parsing the nodes of an AAG into subgraphs. However, this AAG method could only be applied to extract polyhedral objects that have negative convexity. Gavankar and Henderson \citep{gavankar1995graph} used special algorithms to extract the faces that are connected with one or more faces, such as the top surface of a cylindrical protrusion. However, this method can not be applied to notches and protrusions that are linked to more than two faces of another feature.

\begin{figure}
  \centering
  \includegraphics[width=64mm]{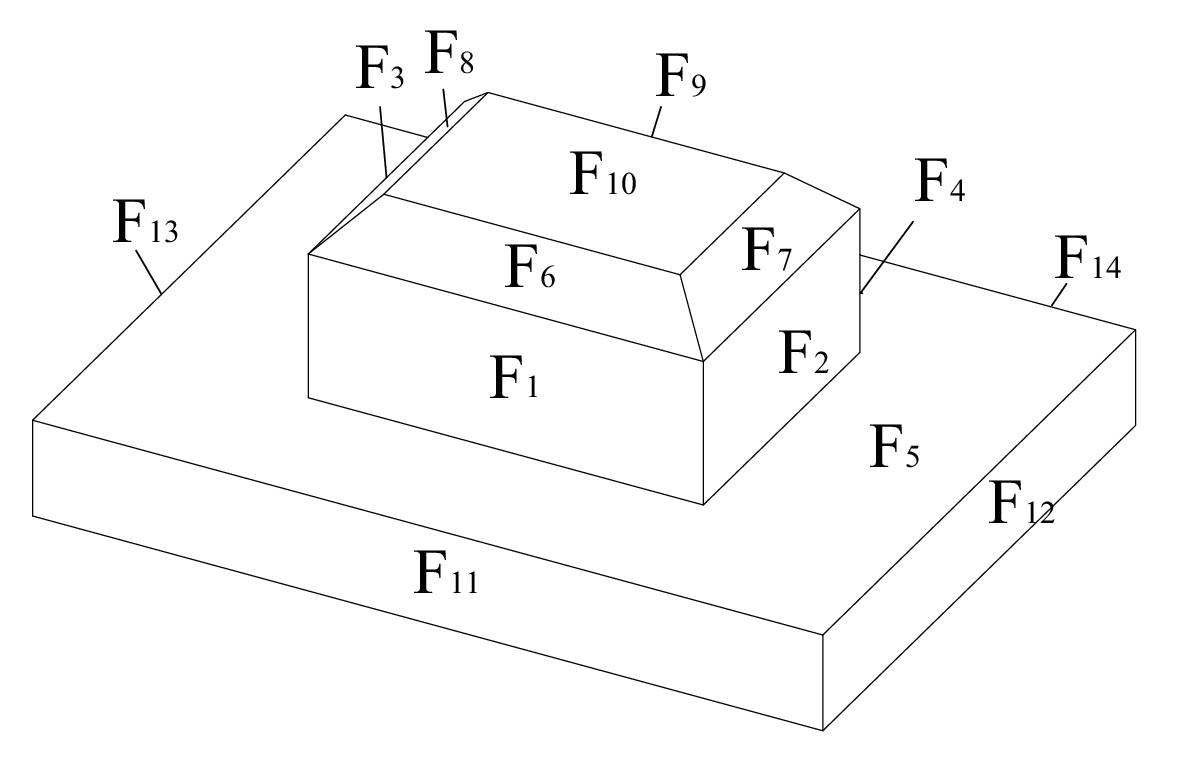}
  \caption{An exception of the MAAG method}
\label{fig:1}       % Give a unique label
\end{figure}

The defects of AAGs were not significantly eliminated until the concept of multi-attributed adjacency graph (MAAG) proposed by Venuvinod and Wong \citep{venuvinod1995graph}. MAAG is built by adding extra attributes to the nodes and arcs of AAG to expand the capabilities of the AAG. However, this method could only extract cylindrical features or features related to curved surfaces, not including polyhedral features. Yuen and Venuvinod~\citep{yuen1999geometric} experimented with the concept of a multi-attributed adjacency matrix (MAAM) and sought for `less expert system and more algorithmic way’ for feature recognition. Two types of complex features, that is, protrusion and depression, could be identified with two recognition algorithms CvTA and CxTA. However, there are two shortcomings of MAAMs. First, all objects must be confined to `three-corner’ solids. Second, protrusion features are limited to simple structures. The reason is that the algorithm CxTA merely covers the root and boundary faces while convex feature faces may contain other faces. For example, there is a chamfer in the protrusions as shown in Fig.~\ref{fig:1}. $\{F_1, F_2, F_3, F_4, F_5\}$ are root faces and $\{F_6, F_7, F_8, F_9, F_{11}, F_{12}, F_{13}, F_{14}, F_{15}\}$ are boundary faces. But the CxTA algorithm only identifies the faces $\{F_1, F_2, F_3, F_4, F_6, F_7, F_8, F_9\}$, not including $F_{10}$. Thus, the protrusion could not be extracted completely. Qamhiyah et al. \citep{qamhiyah1996geometric} proposed a feature extraction technique that was based on loop-adjacency hypergraphs. This technique focused on the generalized class properties of features with planar faces only. Besides, there are some other variations of the AAG method, such as the EAAG method \citep{gao1998automatic}, the IFRM method \citep{nasr2006new}. However, all of them fail to extract protrusion and depression features completely.

\begin{figure}
  \centering
  \includegraphics[width=64mm]{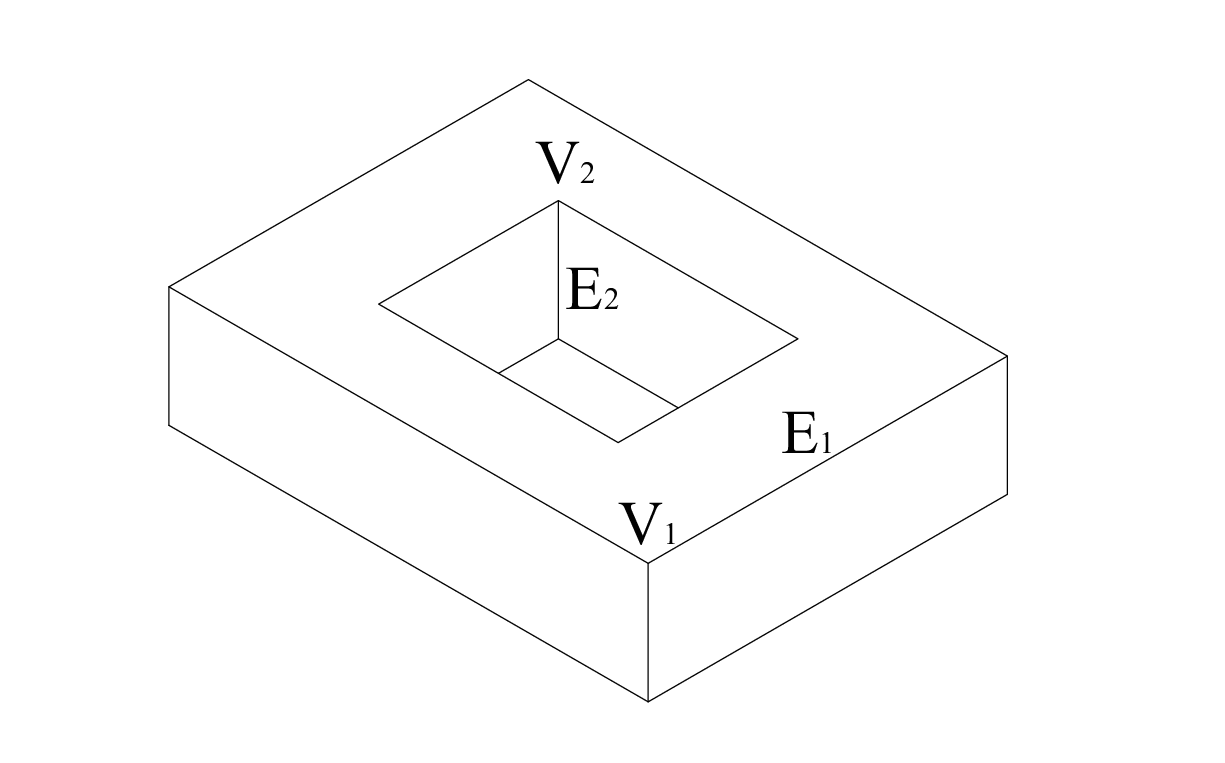}
  \caption{An exception of the neutral basis method}
\label{fig:2}       % Give a unique label
\end{figure}

Zhu et al. \citep{zhu1998neutral} proposed the neutral basis idea and developed the extraction algorithm. They extract from CAD models two types of geometric features, including protrusion and depression features. These two feature types are called neutral bases to facilitate the matching process in a knowledge library. However, their method may result in a disorder between \glspl{feature subgraph}, particularly in the transitory areas. The limitation is attributed to no clear criteria to outline neutral bases in this method. For example, there is a through polyhedral hole in a block as shown in Fig.~\ref{fig:2}. If we start with the vertex $V_1$ with bi-attribute (1,1) in the model, all faces would be connected. As there is no entity with bi-attribute (-1,-1), no concave features could be obtained. This extraction result does not match the through polyhedral hole feature.

Besides the graph-based methods above, many other methods have been proposed. The hint-based method ~\citep{vandenbrande1993spatial,regli1995geometric,regli1995extracting} was developed to find traces of a milling cutter in the part boundary. These traces are used to generate a feature volume. Even though this method succeeded in recognizing interacting features ~\citep{marefat1990geometric}, it is difficult to find correct traces for features. Volumetric-based representation is another category of feature extraction. Woo ~\citep{tang1991algorithmic,tang1991algorithmic2} proposed the Alternating Sum of Volumes (ASV) decomposition for feature extraction. The ASV decomposition represents a non-convex object as a Boolean combination of convex components. Kim and Wilde~\citep{kim1992convex} combined a partitioning procedure with Woo’s ASV decomposition to obtain the ASVP decomposition. Bayesian approaches are proposed to handle uncertainties and find the set of missing arcs. A BN is a probabilistic graphical model that represents a set of random variables and their conditional dependencies via a directed acyclic graph~\citep{cai2017bayesian,xu2022modeling,xu2022mathematical, xu2023ucb}. For example, Ji and Marefat~\citep{ji1995bayesian} used Bayesian networks to restore missing arcs. In addition, some hybrid methods are also proposed. Rahmani and Amirkabir \citep{rahmani2006boundary} proposed a hybrid of graph-based and hint-based methods to recognize interacting features. Rameshbabu and Shunmugam \citep{rameshbabu2009hybrid} developed a hybrid method that involves both volume subtraction and face adjacency graph to handle geometric features. However, all those methods fail to extract complex convex features in the true sense.

\section{Proposed Feature Extraction Approach}
\label{Methodology}

%\subsection{Framework of extraction of neutral feature}
%\label{Framework of extraction of neutral feature}
A set-based feature extraction approach is proposed to extract a set of feature sub-graphs from boundary representation (B-REP) models. First, we define the convexity of all geometric entities based on bi-attribute \citep{zhu1998neutral}. A TAAG model is also defined to integrate the connections and attributes of faces, edges, and vertices of a B-REP. These definitions could bring convenience to the formation of feature subgraphs. Second, a feature extraction workflow is designed in two stages. The first stage is to identify geometric entities with the same convexity as feature boundaries from the point-edge graph edges. The second stage is to build feature subgraphs with faces and edges that share the same convexity in the edge-plane graph.

\subsection{Convexity of geometric entities}
\label{Convexity definition of topological entities}

According to the Global Gauss-Bonnet Theorem \citep{zhu1998neutral,do1976differential}, even though adding one geometric feature to an existing object will lead to variations of curvature distributions, the total curvature remains invariant as long as this deformation does not affect the Euler characteristic. The curvature of an object can be simplified as the convexity of geometric entities for CAD models. Thus, our approach starts with defining the convexity of basic geometric entities, including faces, edges, and vertexes.

\subsubsection{Face}
\label{Convexity of face}
In this study, all faces are assumed to be limited to planes and ruled surfaces which are common in most CAD models. The general type of surfaces is reserved for future work. While each face has two principal curvatures k$_1$ and k$_2$ in two perpendicular directions, the convexity of a face can be calculated with these two curvatures. If one curvature is negative, the face is convex in the corresponding direction. If it is positive, the face is concave. The convexity of a face is determined by both $\{k_1, k_2\}$ and their product k$_1$*k$_2$, as summarized in Table ~\ref{table1:Convexity of face}. A convex face is denoted as +1 and a concave face is -1. A transitory face is denoted as 0.

\begin{table}
\caption{Convexity of face}
\label{table1:Convexity of face}       % Give a unique label
  \centering
\begin{tabular}{cccc}
\hline\noalign{\smallskip}
k$_1$ & k$_1$ & k$_1$*k$_2$ & Results \\
\noalign{\smallskip}\hline\noalign{\smallskip}
$<$0 & $\leqslant$0 & $\geqslant$0 & convex face \\
$\leqslant$0 & $<$0 & $\geqslant$0 & convex face \\
$>$0 & $\geqslant$0 & $>$0 & concave face \\
$\geqslant$0 & $>$0 & $>$0 & concave face \\
$=$0 & $=$0 & $=$0 & transitory face (plane) \\
  &   & $<$0 & transitory face \\
\noalign{\smallskip}\hline
\end{tabular}
\end{table}

\subsubsection{Edge}
\label{Convexity of edge}
\begin{figure}
  \centering
  \includegraphics[width=64mm]{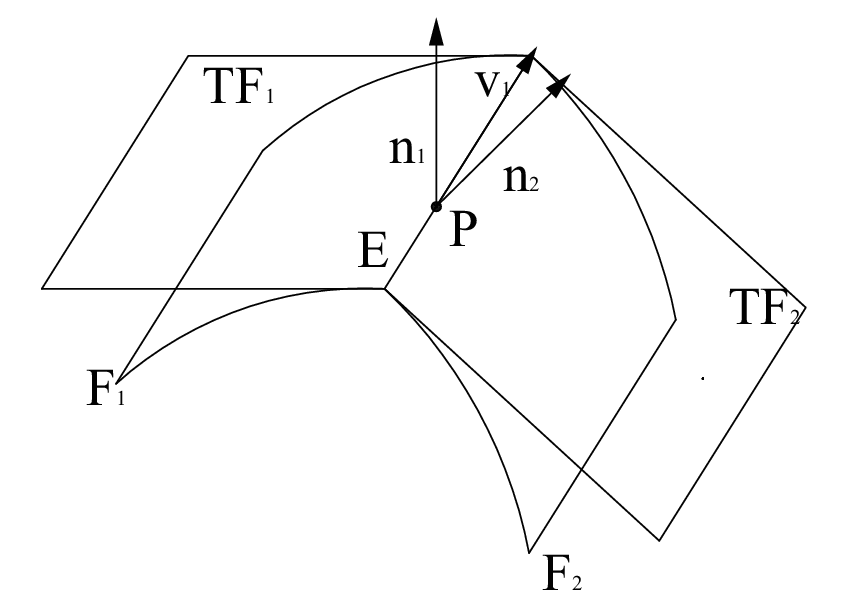}
  \caption{Convexity of edge}
\label{fig:4}       % Give a unique label
\end{figure}

The convexity of an edge is determined by the intersection angle between two faces adjacent to the edge. As shown in Fig.~\ref{fig:4}, the target edge is $E$ and it has two adjacent faces $F_1$ and $F_2$. Let \textbf{$n_1$} and \textbf{$n_2$} be the normal vectors of the tangent planes $TF_1$ and $TF_2$ of the faces $F_1$ and $F_2$ at some point $P$ on the edge. Then, the convexity of the edge could be computed according to Eq.~\ref{eq:1}. A convex edge is marked as +1, a concave edge is -1, and a transitory edge is 0.

\begin{equation} \label{eq:1}
b_1 =\left\{
\begin{aligned}
+1&(convex) \\
 0&(transitory)\\
-1&(concave)
\end{aligned}
\right.
,if\ \mathbf{n_1} \times \mathbf{n_2} \cdot \mathbf{v_1} \left\{
\begin{aligned}
&>&0 \\
&=&0 \\
&<&0
\end{aligned}
\right.
\end{equation}

\begin{figure}
  \centering
  \includegraphics[width=72mm]{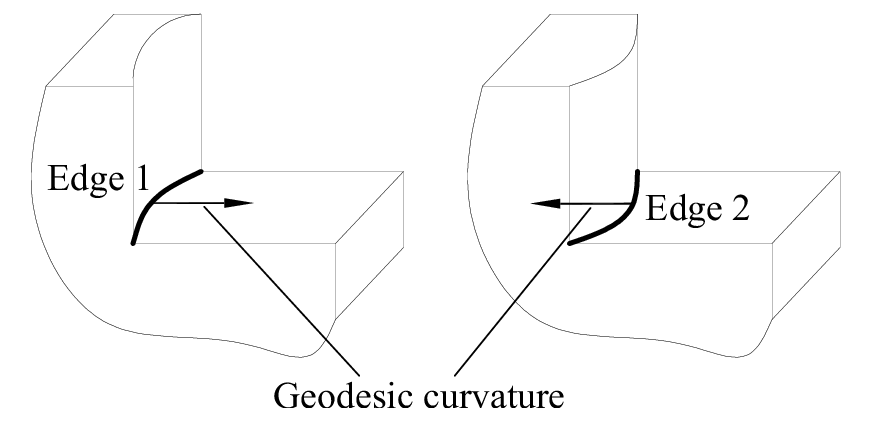}
  \caption{Geodesic curvature of edge}
\label{fig:5}       % Give a unique label
\end{figure}

It is worth mentioning that the geodesic curvature of an edge is not considered in Eq.~\ref{eq:1}. For example, Edge 2 is concave intuitively even though its geodesic curvature is convex, as shown in Fig.~\ref{fig:5}. If the geodesic curvature was considered a determining factor, Edge 2 would become transitory, which contradicts the intuition. So, for simplicity, we only leverage the intersection angle to determine the convexity.

\subsubsection{Vertex}
\label{Convexity of vertex}
A simple count method is adopted in this study to determine the convexity of a vertex for efficiency. As to a target vertex, all adjacent edges are listed, and their convexity is determined with the rules above. If all adjacent edges are not concave, the vertex is convex and marked as 1. If all adjacent edges are not convex, the vertex is concave and marked as -1. In the rest cases, the vertex is transitory and marked as 0.

\subsection{Two-level Attribute Adjacent Graph}
\label{Two-level Attribute Adjacent Graph}

In previous studies, an AAG and its extended graph models only contain the connection and attributes of faces and edges. This shortcoming makes complex features hard to extract. The reason is that faces and edges may be insufficient to separate two features. As feature boundaries consist of vertexes and edges, vertex-edge graphs should also be considered in feature extraction. Thus, a TAAG model is proposed to combine both vertex-edge graphs and edge-face graphs as a whole while they are built independently. This TAAG model is defined as follows: 

\begin{enumerate}
  \item A TAAG model has two components, including a vertex-edge graph and an edge-face graph. The vertex-edge graph is a network between all vertexes and edges. The edge-face graph is a network between edges and faces.
  \item In a vertex-edge graph, a vertex $V_i$ of the part is shown as a unique node $P_{i}$. In an edge-face graph, a face $F_k$ of the part corresponds to a corresponding node $P_{k}$.
  \item If two faces $F_i$, $F_j$ share the same edge, there is a unique link $A^f_{ij}$ in the edge-face graph. If two vertices $V_i$ and $V_j$ share the same edge, there is a corresponding link $A^v_{ij}$ in the vertex-edge graph.
  \item All elements of a TAAG have a convexity attribute according to the corresponding convexity of the CAD model. The attribute value is assigned as -1 for a concave element, +1 for a convex element, and 0 for a transitional element.
\end{enumerate}

\begin{table*}
\caption{TAAG}
\label{TAAG Table}       % Give a unique label
  \centering
\begin{tabular}{cccc}
\hline\noalign{\smallskip}
Name & Model & Vertex-edge Graph & Edge-face Graph \\
\noalign{\smallskip}\hline\noalign{\smallskip}
Protrusion & \includegraphics[scale=0.8]{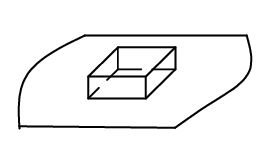} & \includegraphics[scale=0.8]{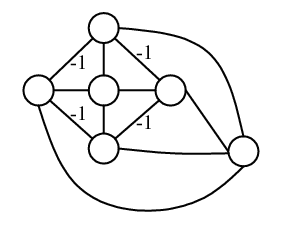} & \includegraphics[scale=0.8]{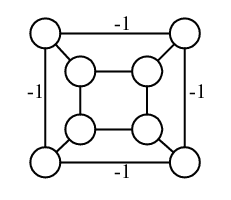} \\
Step & \includegraphics[scale=0.8]{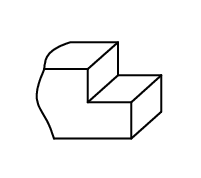} & \includegraphics[scale=0.8]{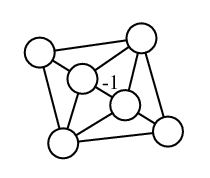} & \includegraphics[scale=0.8]{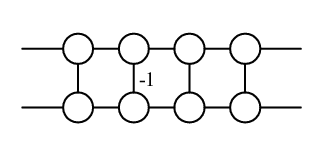} \\
\noalign{\smallskip}\hline
\end{tabular}
\end{table*}

The difference between TAAG and previous graph models is the two-layer structure. For example, a protrusion feature and a step feature can be built in the TAAG form, as shown in Table~\ref{TAAG Table}. The modeling procedure of a TAAG is designed as follows:

\begin{enumerate}
  \item  Obtain all the vertices, edges, and faces and determine their convexity;
  \item  Initialize all nodes in a TAAG. For each vertex, there is a node in the vertex-edge graph. For each face, there is a node representing the face in the edge-face graph.
  \item  Traverse all the edges in the model and find the adjacent faces and vertices of them. If there is an edge between two vertices or faces, add a link between the corresponding nodes in the graph.
  \item  Assign an attribute value to each graph element according to the convexity of the geometric entity.
\end{enumerate}

\subsection{Feature extraction workflow}
\label{Feature extraction process of neutral feature}
\begin{figure*}[tbp]
  \centering
  \includegraphics[width=0.85\textwidth]{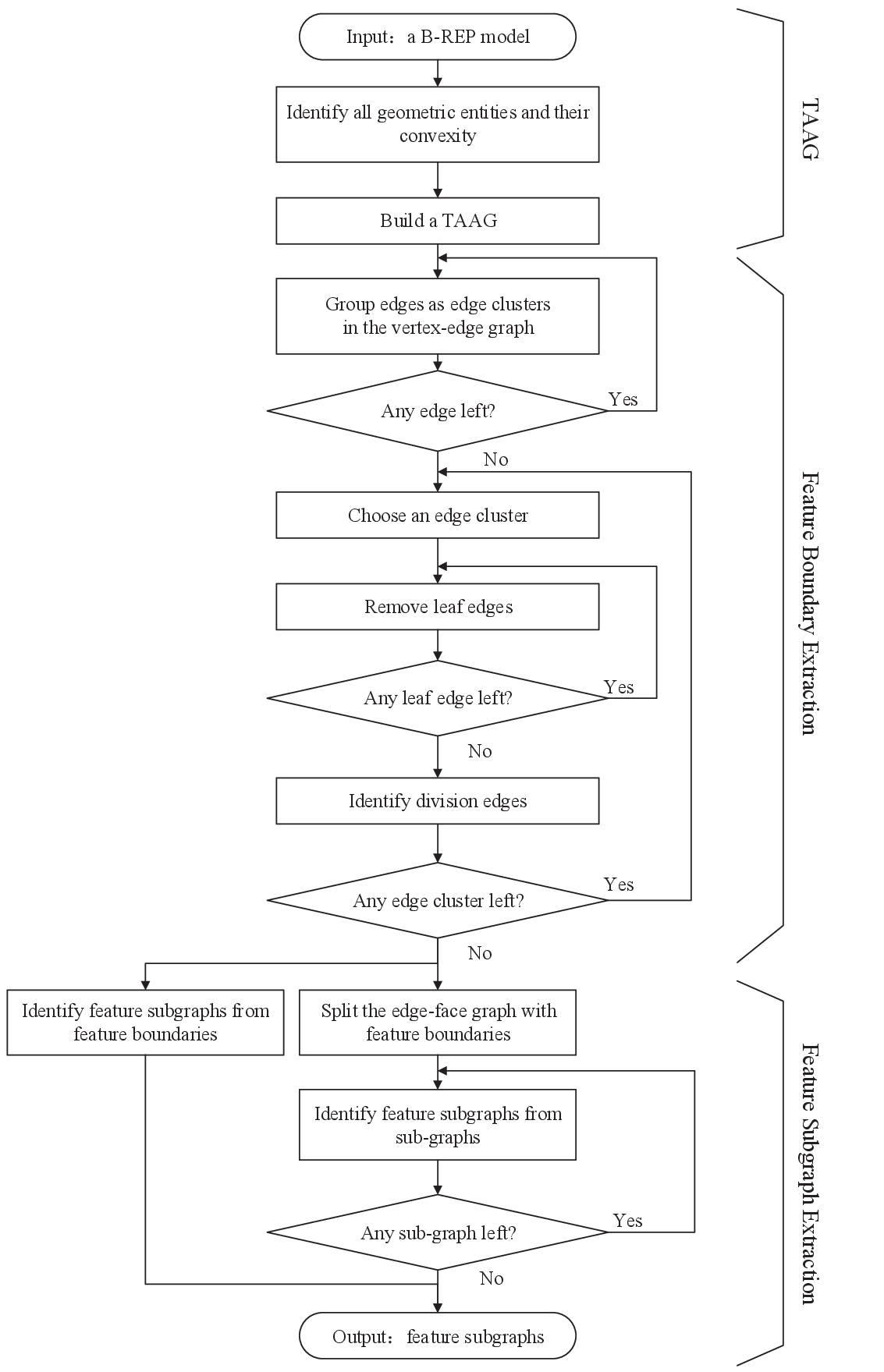}
  \caption{Feature extraction workflow}
\label{flowchart}       % Give a unique label
\end{figure*}

The workflow of the proposed feature extraction approach is shown in Fig.~\ref{flowchart}. It consists of three stages. First, the convexity (convex, concave, and transitory) of all geometric entities is identified to build a TAAG model. Second, the edges in a vertex-edge graph are connected into different \glspl{edge cluster}. Then, we refine the structure of each edge cluster into some \glspl{edge loop}. A \gls{feature boundary} is determined from an edge loop by checking if each edge is a \gls{division edge}. Third, the edge-face graph is partitioned into sub-graphs using the feature boundaries. All sub-graphs along with feature boundaries are used to identify feature subgraphs. The second and third stages are realized by two algorithms in the following sections. Two CAD models are designed as examples to demonstrate the algorithms, as shown in Fig.~\ref{Program1} (a),(b).

\subsubsection{Feature boundary extraction algorithm}
\label{Feature boundary extraction algorithm}

This algorithm is designed to extract feature boundaries in four steps.

\begin{figure*}
  \centering
  \includegraphics[width=0.9\textwidth]{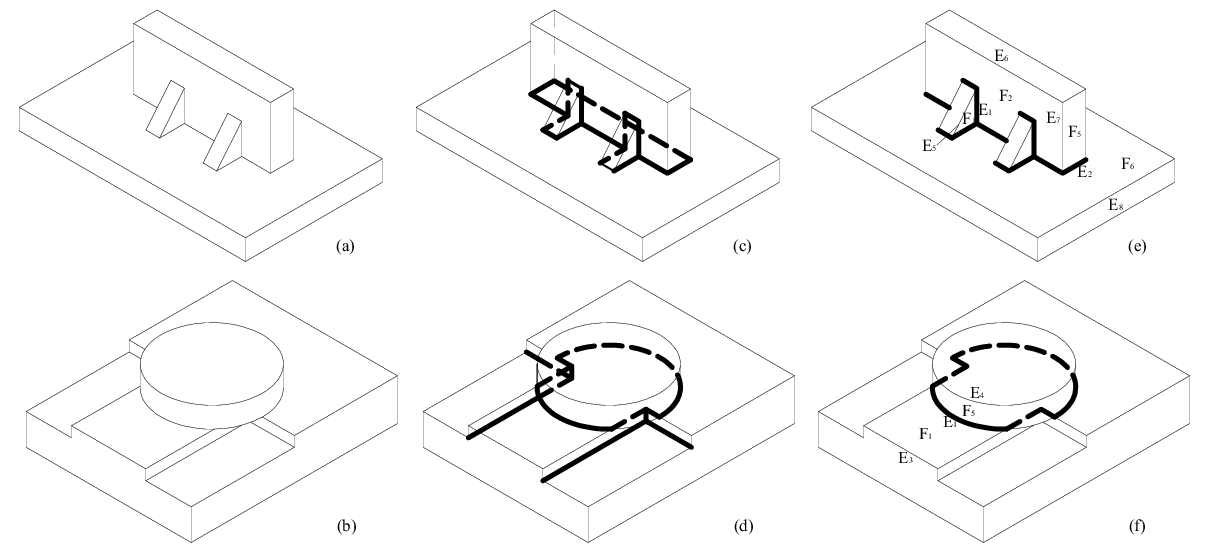}
  \caption{Feature boundary extraction}
\label{Program1}       % Give a unique label
\end{figure*}

\begin{enumerate}
    \item Input a TAAG of a CAD model.
    \item Group all adjacent edges that share the same convexity characteristic into an edge cluster in the vertex-edge graph. The breadth search algorithm is used to find all edge clusters.

    In Case A and B, the concave edge clusters are shown as the bold lines in Fig.~\ref{Program1} (c),(d).
    
    \item Find the edge loop in each edge cluster. This step is realized by removing all \glspl{leaf edge} from an edge cluster with iterations. Thus, all remaining edges constitute an edge loop in this edge cluster.

    In Case A, the concave edge cluster contains no leaf edges. So, all edges constitute an edge loop directly, as shown in Fig.~\ref{Program1} (c).

    In Case B, there are six leaf edges in the concave edge cluster. After removing them, we have an edge loop that contains six edges, as shown in Fig.~\ref{fig:7} and Fig.~\ref{Program1} (f).

\begin{figure}
  \centering
  \includegraphics[width=0.75\textwidth]{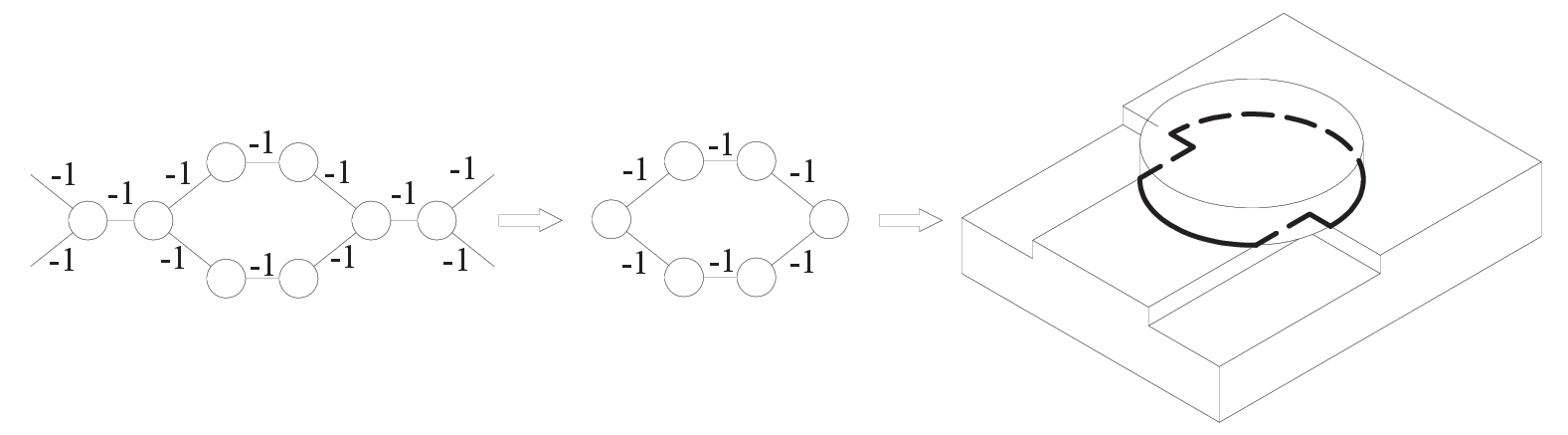}
  \caption{Edge loop and feature boundary}
\label{fig:7}       % Give a unique label
\end{figure}

    \item Check whether the edges in the edge loop are division edges in the edge-face graph. If an edge is a division edge, it can be used as a feature boundary.

    In Case A, all edges in the edge loop are division edges. For example, as to the edge $E_1$, there are two convex edges $E_5$ and $E_6$ on two sides of $E_1$ separately. So, $E_1$ is a division edge, as shown in Fig.~\ref{Program1}(e). Thus, the whole edge cluster is a feature boundary.
    
    In Case B, all edges in the edge loop are division edges. Taking the edge $E_1$ as an example, there are two edges $E_3$ and $E_4$ on two sides of $E_1$, as shown in Fig.~\ref{Program1} (f). Thus, the six edges of the edge loop can be used as a feature boundary.
\end{enumerate}

\subsubsection{Feature subgraph extraction algorithm}
\label{Feature subgraph extraction}

\begin{figure*}[htb]
  \centering
\subfigure[Case A]{
\begin{minipage}[c]{0.48\textwidth}
\centering
  \includegraphics[width=75mm]{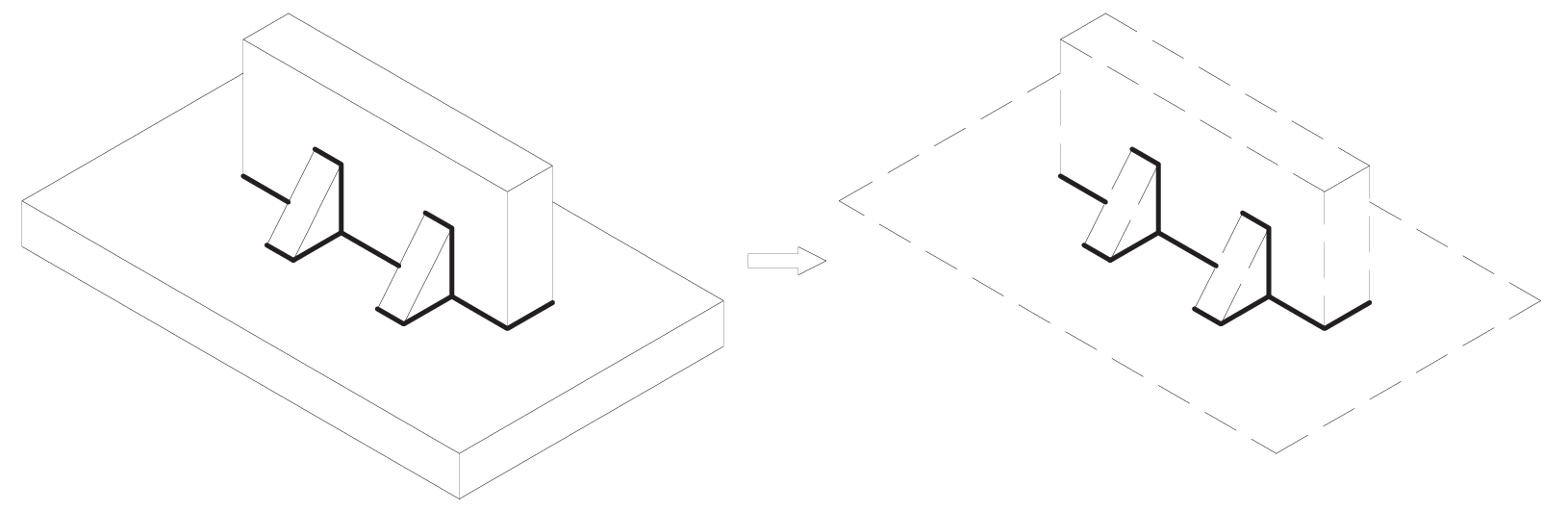}
\end{minipage}%
}%
\subfigure[Case B]{
\begin{minipage}[c]{0.48\textwidth}
\centering
  \includegraphics[width=75mm]{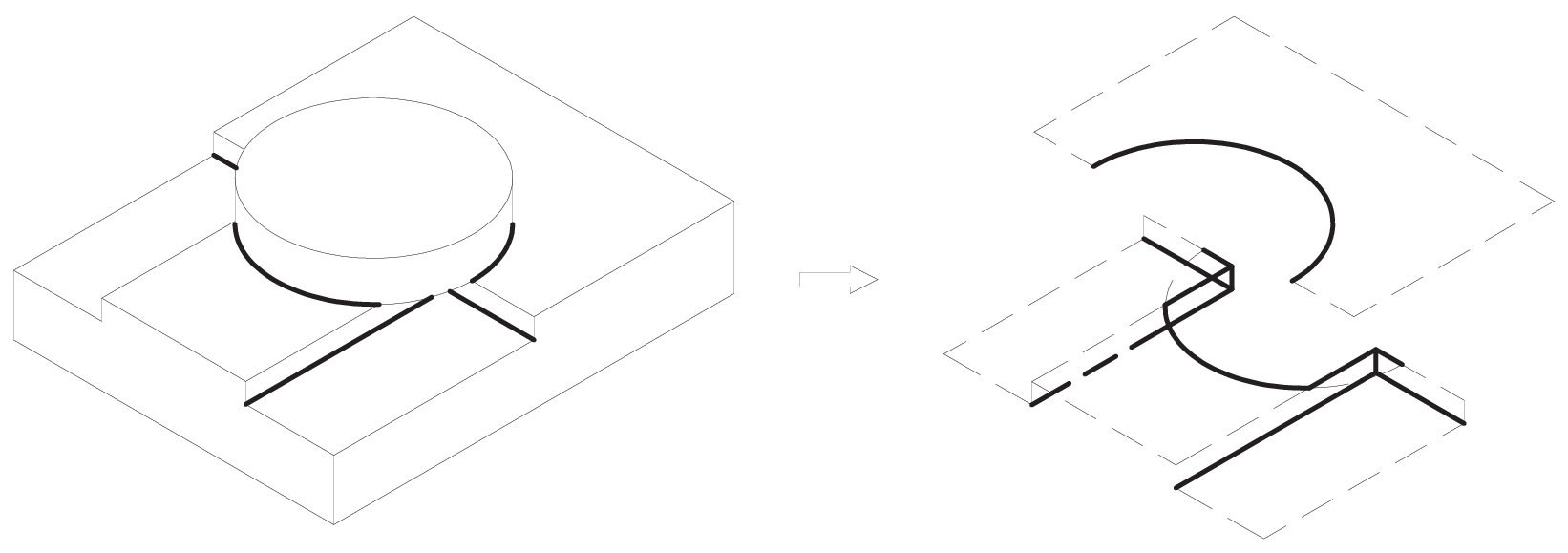}
\end{minipage}
}
\caption{Identify geometric entities adjacent to feature boundaries}\label{fig:21}
\end{figure*}

\begin{figure*}[htbp]
  \centering
\subfigure[Case A]{
\begin{minipage}[c]{0.48\textwidth}
\centering
  \includegraphics[width=75mm]{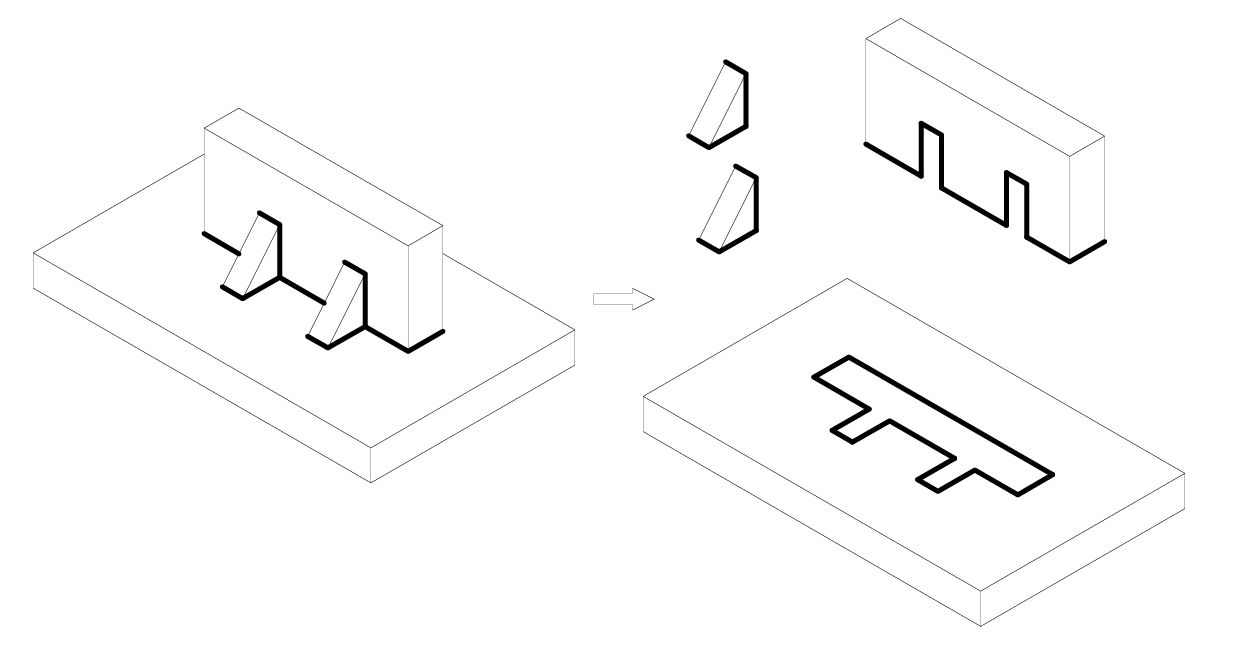}
\end{minipage}%
}%
\subfigure[Case B]{
\begin{minipage}[c]{0.48\textwidth}
\centering
  \includegraphics[width=75mm]{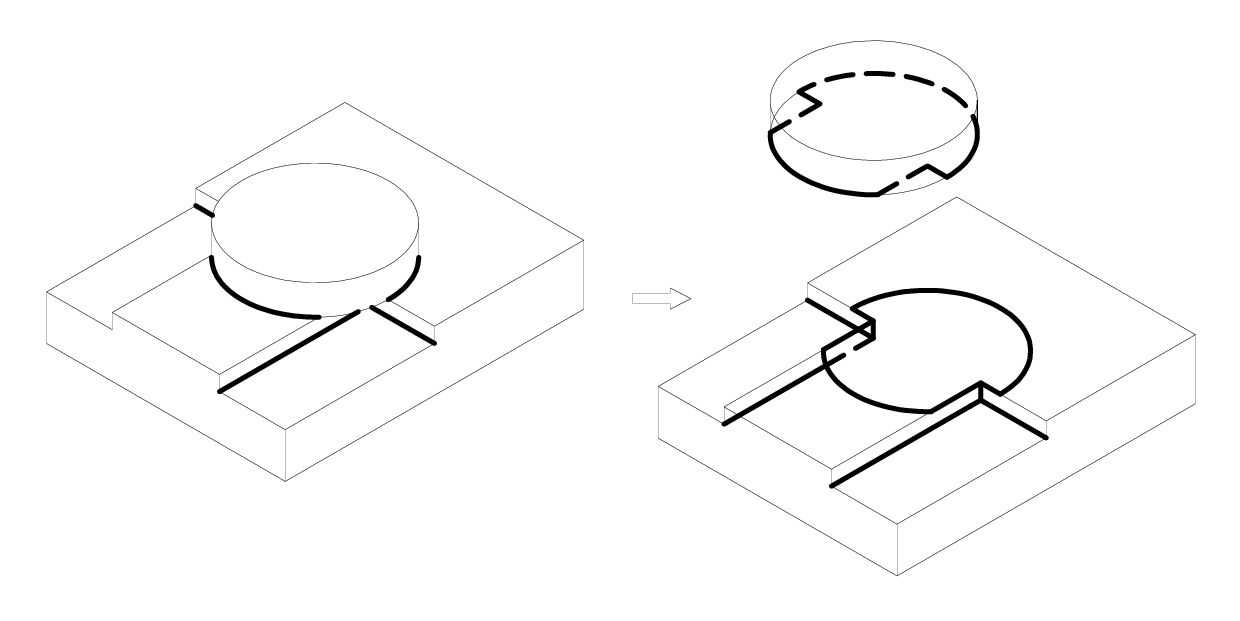}
\end{minipage}
}
\caption{Break edge-face graphs into subgraphs}\label{fig:8}
\end{figure*}

\begin{figure*}[htbp]
  \centering
\subfigure[Case A]{
\begin{minipage}[c]{0.48\textwidth}
\centering
  \includegraphics[width=75mm]{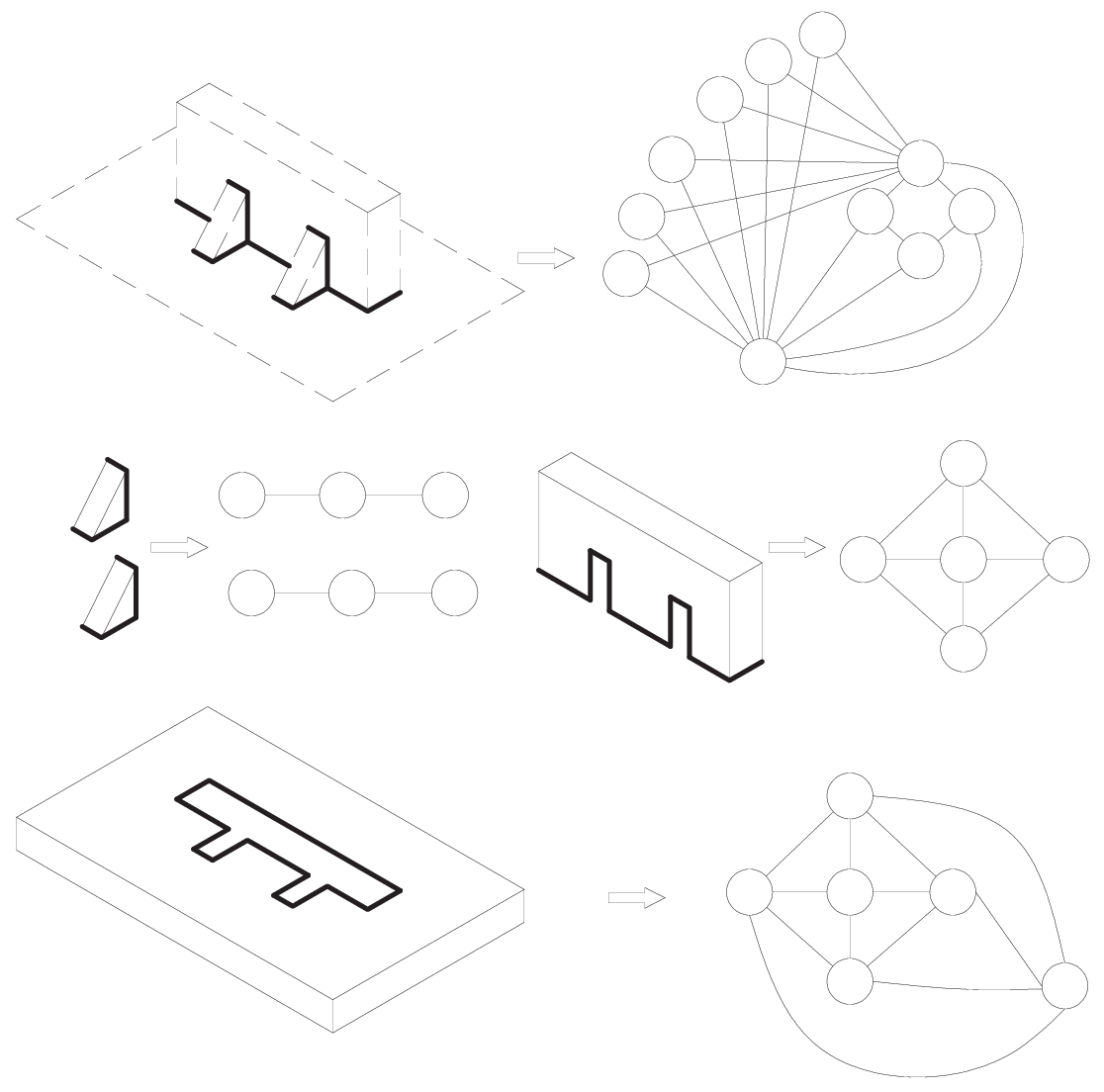}
\end{minipage}%
}%
\subfigure[Case B]{
\begin{minipage}[c]{0.48\textwidth}
\centering
  \includegraphics[width=75mm]{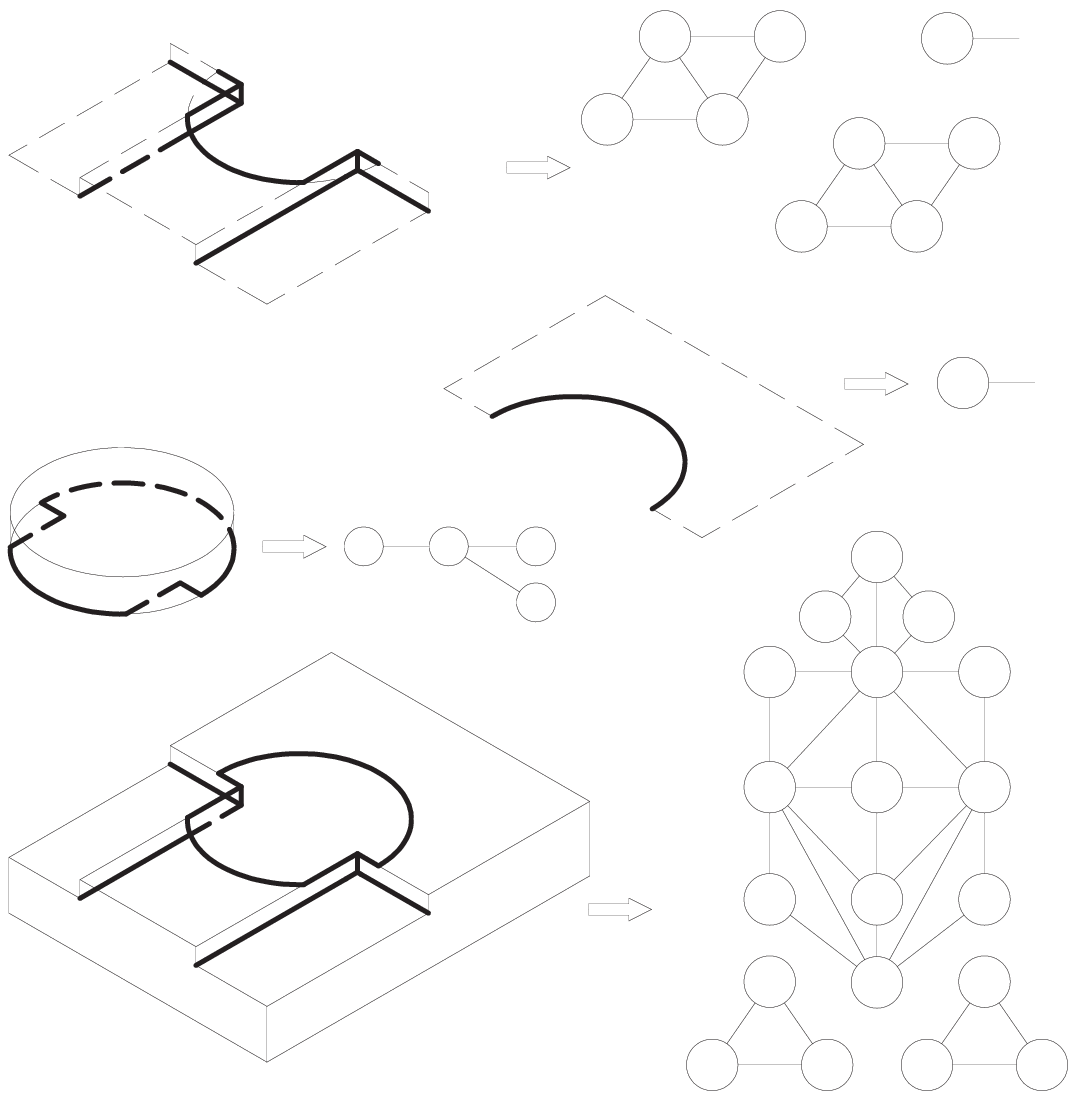}
\end{minipage}
}
\caption{Feature subgraphs}\label{fig:9}
\end{figure*}

Given the generated feature boundaries, this algorithm is designed to extract a set of feature subgraphs with the same convexity in four steps.

\begin{enumerate}
    \item Build feature subgraphs with the feature boundaries identified in Section~\ref{Feature boundary extraction algorithm} by connecting a feature boundary with all adjacent geometric entities in the edge-face graph. All connected geometric entities should be transitory or have the same convexity.
    
    In Case A, there are eleven planes adjacent to the concave edge cluster, as shown in Fig.~\ref{fig:21} (a). So, the feature subgraph of these eleven planes is the complementary space of the block and two triangular prisms.

    In Case B, there are eight planes adjacent to the concave edge cluster, as shown in Fig.~\ref{fig:21} (b). The corresponding feature subgraph consists of two step-shape spaces.

    \item Break the edge-face graph into sub-graphs with identified feature boundaries. That is, all division edges are removed from the edge-face graph and the edge-face graph is divided into a set of subgraphs. All subgraphs should not be connected with each other because of the feature boundaries. It should be noted that leaf edges in an edge cluster are not removed because they are not part of a feature boundary.
    
    In Case A, after removing the feature boundaries, the edge-face graph is broken into four sub-graphs as shown in Fig.~\ref{fig:8} (a).
    
    In Case B, after removing the feature boundaries, there are two sub-graphs, as shown in Fig.~\ref{fig:8} (b).

    \item Extract a set of feature subgraphs from all the sub-graphs. Each pair of adjacent entities in a subgraph is compared concerning their convexity. If they share the same convexity, they are part of the same feature subgraph. This comparison is iterated until all geometric entities are classified into feature subgraphs. A convex feature subgraph contains convex or transitory geometric entities, while a concave feature subgraph contains concave or transitory geometric entities.

    In Case A, all geometric entities of the five sub-graphs have the same convexity. So, all five sub-graphs are feature subgraphs, as shown in Fig.~\ref{fig:9} (a). 
        
    In Case B, eight feature subgraphs could be identified from the four sub-graphs, as shown in Fig.~\ref{fig:9} (b). 

    \item Identify features with feature subgraphs. Each feature subgraph will be recognized as some feature with feature pattern knowledge. A feature pattern library is used in this report to recognize features. For example, the convex feature subgraph with three faces in Fig.~\ref{fig:9} (a) could be recognized as a supporting rod. The development of more feature recognition methods is reserved for future works. It is worth mentioning that not all feature subgraphs have corresponding features. Different domain knowledge may also lead to diverse sets of features.

\end{enumerate}

\section{Implementation}
\label{Implementation}

Current CAD/CAM systems mainly use B-REP models to present solid models. Each CAD/CAM system may have its file format based on B-REP models, such as IGES and STEP. A B-REP model only consists of basic geometric entities, including vertices, edges, and faces. Thus, the input files of our feature extraction system are assumed to be B-REP models. The system is programmed with C++ and UG/Open in the environment of UG NX8.0. Two CAD models with different sets of features are created to test our proposed approach.

\begin{figure*}[htb]
  \centering
\subfigure[Original model]{\label{fig:fft:a1}
\begin{minipage}[c]{0.48\textwidth}
\centering
  \includegraphics[width=75mm]{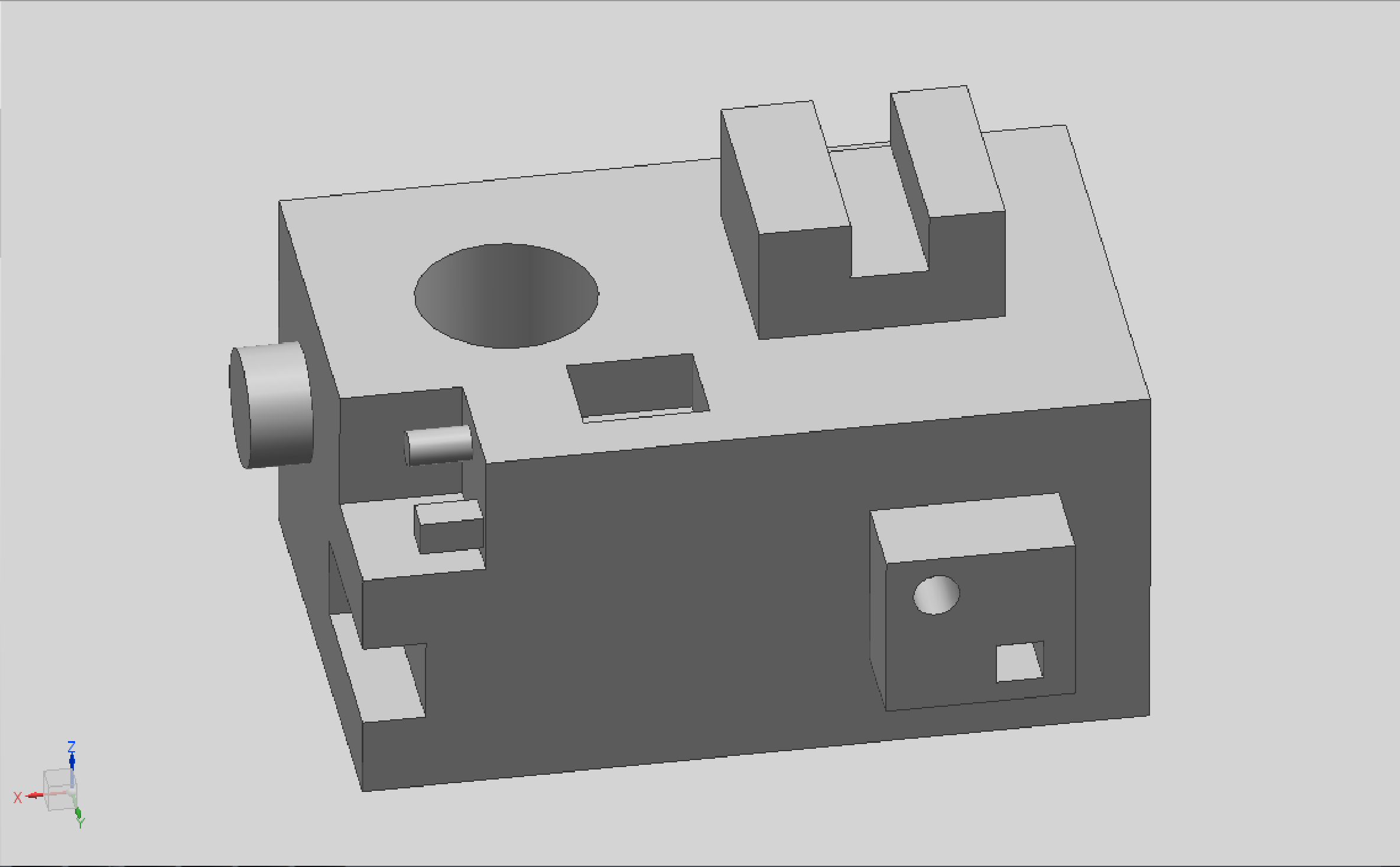}
\end{minipage}%
}%
\subfigure[Extraction results]{
\begin{minipage}[c]{0.48\textwidth}
\centering
  \includegraphics[width=75mm]{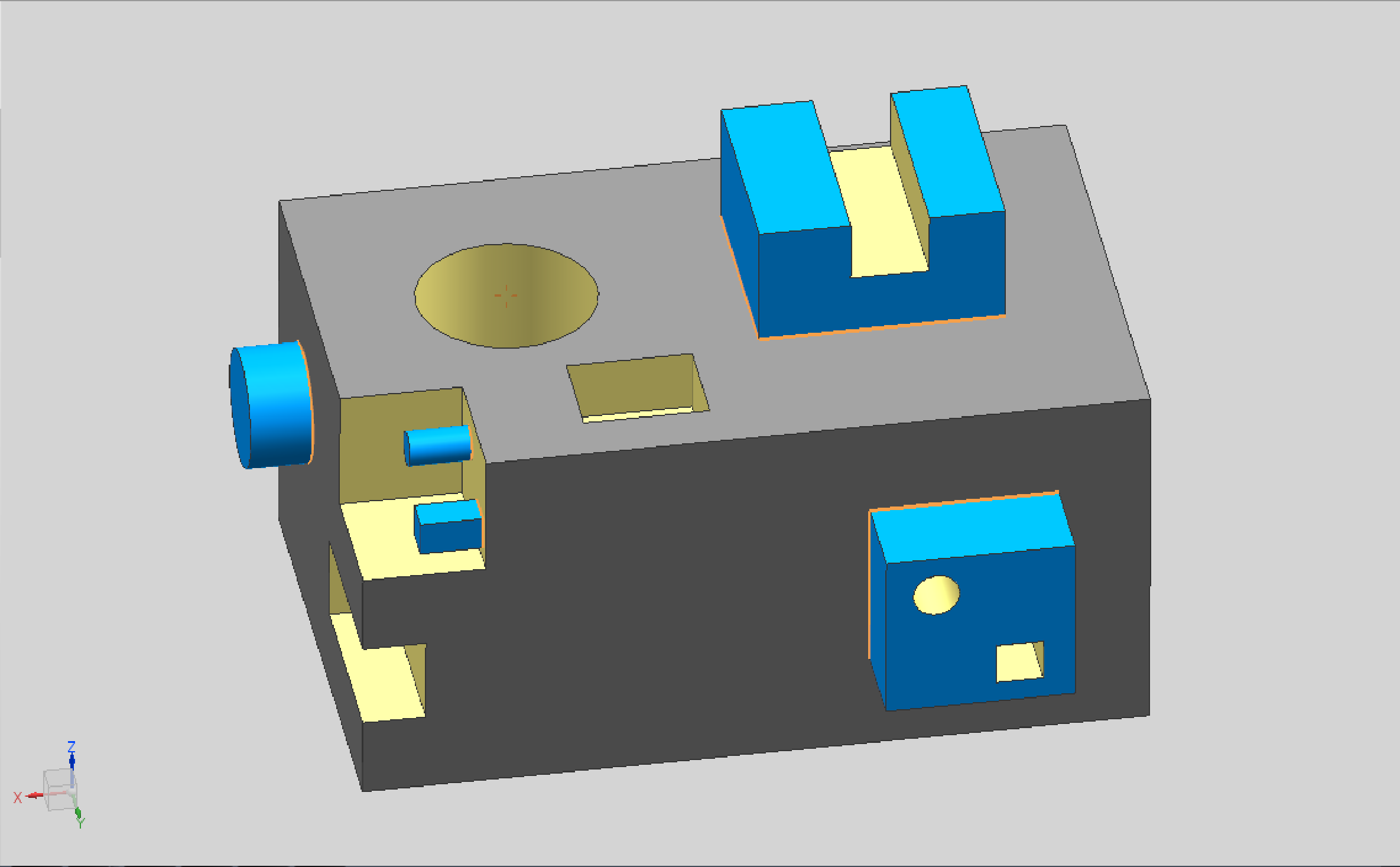}
\end{minipage}
}
\caption{Part 1 and extraction results}\label{fig:fft1}
\end{figure*}

\begin{figure*}[htb]
  \centering
\subfigure[Original model]{\label{fig:fft:a2}
\begin{minipage}[c]{0.48\textwidth}
\centering
  \includegraphics[width=75mm]{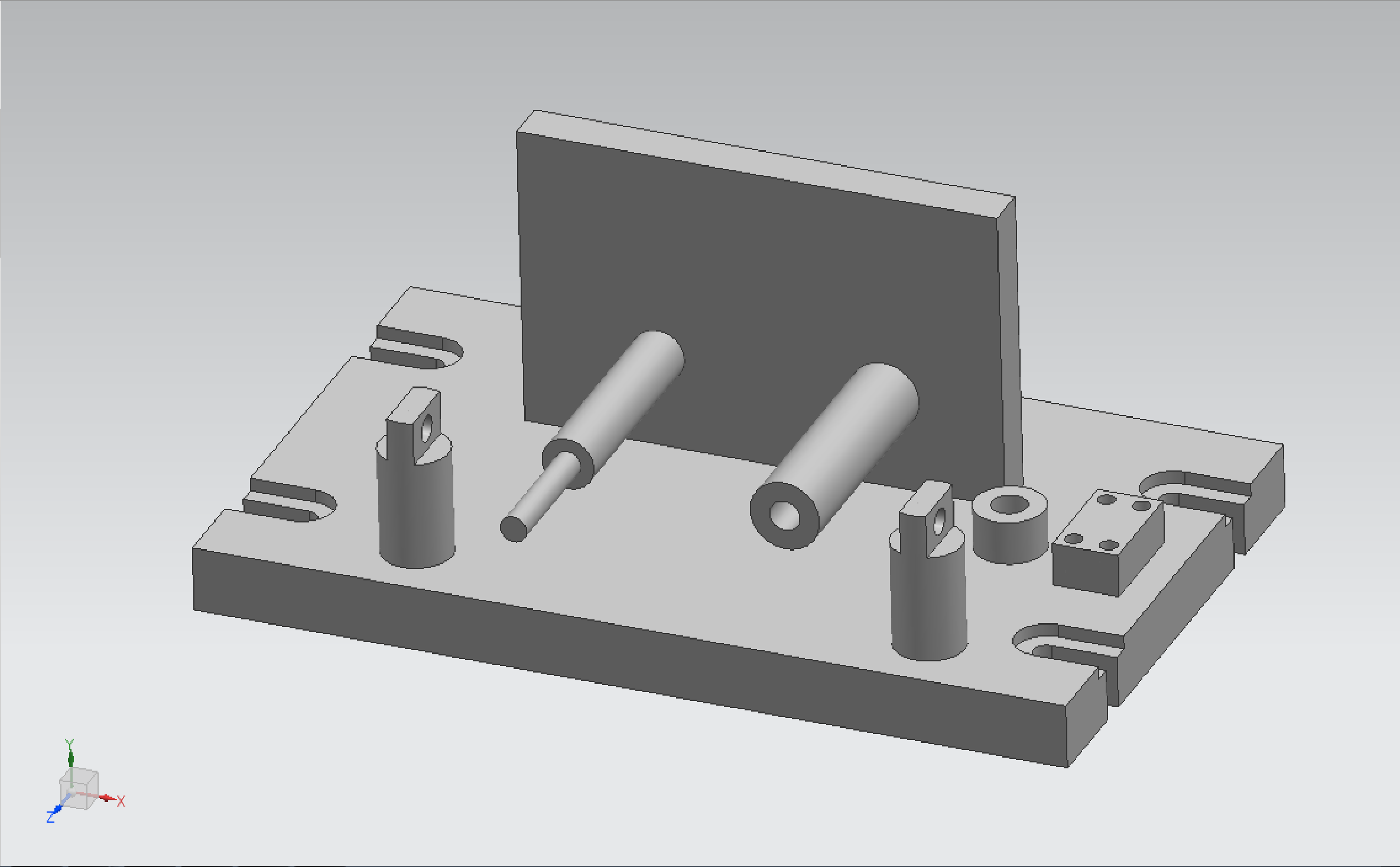}
\end{minipage}%
}%
\subfigure[Extraction results]{
\begin{minipage}[c]{0.48\textwidth}
\centering
  \includegraphics[width=75mm]{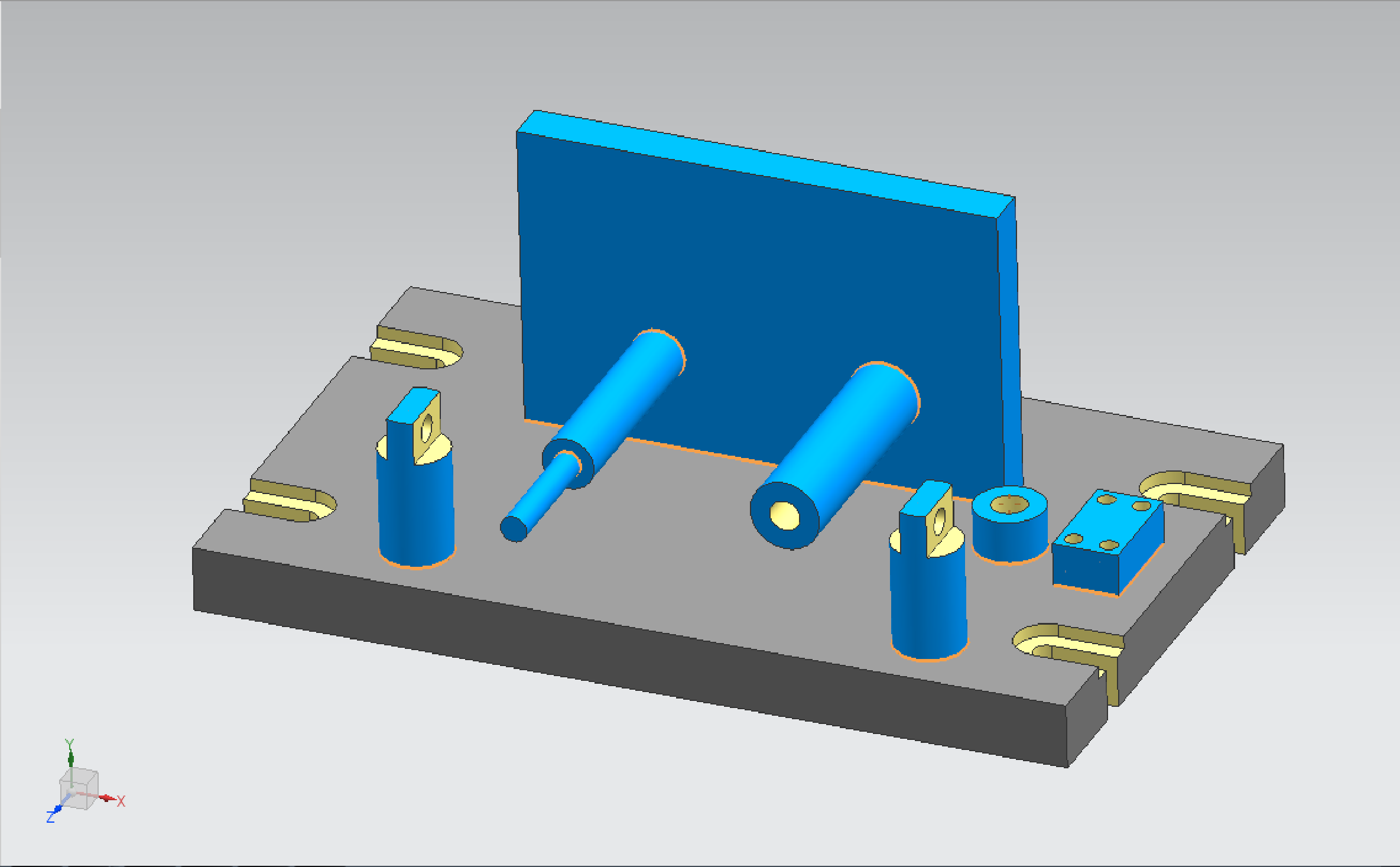}
\end{minipage}
}
\caption{Part 2 and extraction results}\label{fig:fft2}
\end{figure*}

Part 1 is designed with some basic features (boss, holes, blind hole, slot, step, pocket, protrusion). All these basic features are common in mechanical design and analysis. After implementing the developed program, we identify all feature boundaries in the vertex-edge graph. Then, the feature subgraphs of those basic features are extracted in the edge-face graph. The convex feature subgraphs are shown in blue while the concave feature subgraphs are shown in yellow, as shown in Fig.~\ref{fig:fft1}. The feature boundaries are highlighted with red lines. It can be found that all basic features are extracted from the part correctly.

Part 2 is an ordinary fixture part used to fix other mechanical parts. All feature subgraphs and feature boundaries are extracted and displayed in the same way as Part One. The extraction results are consistent with the initial design intents, as shown in Fig.~\ref{fig:fft2}.

\section{Conclusion}
\label{Conclusion}
This report presents a set-based feature extraction approach to extract a set of feature subgraphs from B-REP geometric models. We define the convexity of geometric entities and propose a TAAG model. Then, a feature extraction workflow is designed to identify feature boundaries and feature subgraphs. Our approach can be used to extract both concave and convex feature subgraphs correctly. The authors have tested the approach with some example CAD models to prove its feasibility. As geometric information of B-REP models has inherent uncertainty, it is hard to recognize features with only geometric information. Our approach transforms the uncertainty of geometric information into a set of feature subgraphs. So, the proposed approach offers full flexibility for further feature recognition.

This report also opens some future research directions. First, we mainly extract some simple feature subgraphs from CAD models in this report. In reality, a feature subgraph may contain more complex geometric entities. How to further decompose complex feature subgraphs, especially for interacting features, remains unsolved. Other types of CAD features may be considered to enhance extraction performance. Second, the proposed approach in this report is a preprocessing step for feature recognition. That is, feature recognition methods are still needed to recognize feature subgraphs as features with domain knowledge. Some advanced feature recognition methods could be designed to meet the requirements of feature recognition. Third, more efforts could be devoted to developing artificial intelligence methods to integrate feature extraction and recognition as end-to-end feature recognition systems.

%\printglossary[title=Special Terms, toctitle=List of terms]
\printglossaries

\begin{figure*}[htbp]
  \centering
  \includegraphics[width=0.9\textwidth]{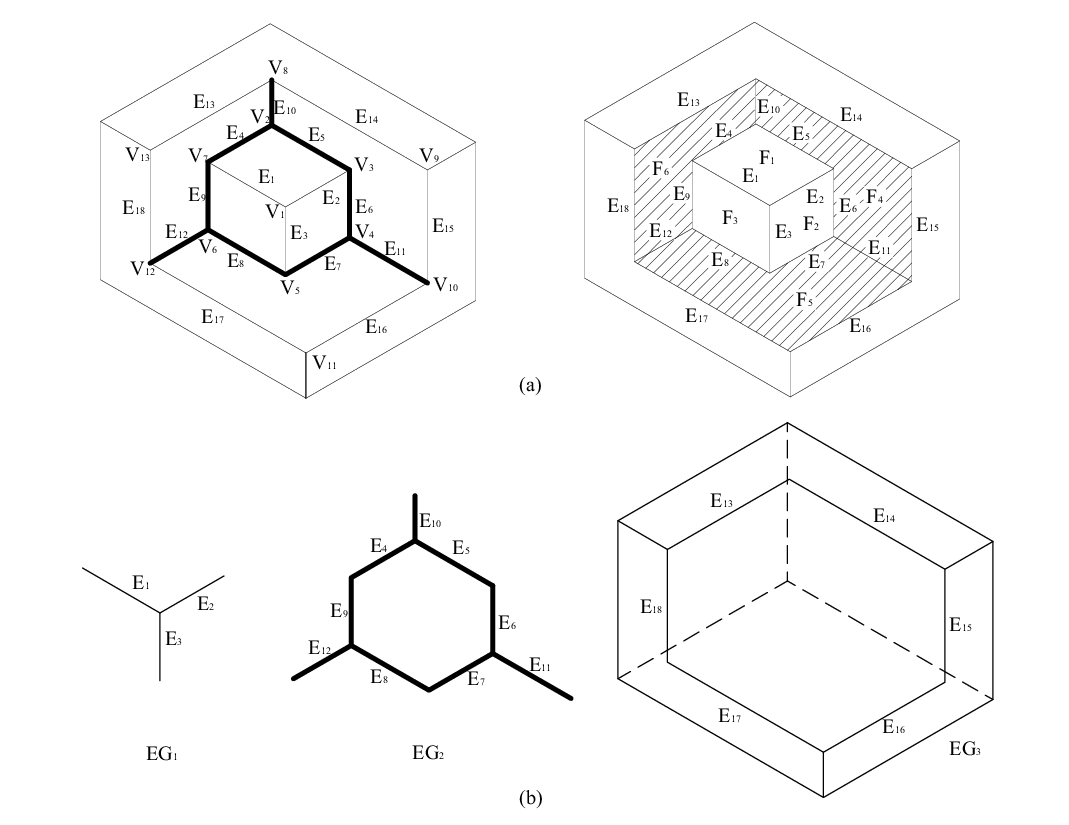}
  \caption{A CAD model for glossaries}
\label{Fig.6}       % Give a unique label
\end{figure*}

%\begin{acknowledgements}
%\end{acknowledgements}


\bibliographystyle{unsrtnat}
\bibliography{references}  %%% Uncomment this line and comment out the ``thebibliography'' section below to use the external .bib file (using bibtex) .

\end{document}